\documentclass[lettersize,journal]{IEEEtran}
\usepackage{hyperref}       
\usepackage{url}            
\usepackage{booktabs}       
\usepackage{amsfonts}       
\usepackage{nicefrac}       
\usepackage{microtype}      
\usepackage{xcolor}         
\usepackage{amsmath}
\usepackage{multirow}
\usepackage{algorithm}
\usepackage{algpseudocode}
\usepackage{graphicx}
\usepackage{amssymb}
\usepackage{caption}

\def\BibTeX{{\rm B\kern-.05em{\sc i\kern-.025em b}\kern-.08em
    T\kern-.1667em\lower.7ex\hbox{E}\kern-.125emX}}
\usepackage{balance}

\begin{document}
\title{Modeling Hierarchical Structural Distance for Unsupervised Domain Adaptation}
\author{Yingxue Xu,
        Guihua Wen,
        Yang Hu,
        and Pei Yang*
\thanks{Yingxue Xu is with South China University of Technology and the Hong Kong University of Science and Technology, email: innse76@outlook.com.}
\thanks{Guihua Wen and Pei Yang are with South China University of Technology, China, e-mail: crghwen@scut.edu.cn and yangpei@scut.edu.cn.}
\thanks{Yang Hu is with University of Oxford, Oxford, UK, email: yang.hu@ndm.ox.ac.uk.}
\thanks{Pei Yang is Corresponding Author.}
\thanks{Copyright © 2024 IEEE. Personal use of this material is permitted. However, permission to use this material for any other purposes must be obtained from the IEEE by sending an email to pubs-permissions@ieee.org.}}
\markboth{Journal of \LaTeX\ Class Files,~Vol.~18, No.~9, September~2020}%
{How to Use the IEEEtran \LaTeX \ Templates}

\maketitle

\begin{abstract}
Unsupervised domain adaptation (UDA) aims to estimate a transferable model for unlabeled target domains by exploiting labeled source data. Optimal Transport (OT) based methods have recently been proven to be a promising solution for UDA with a solid theoretical foundation and competitive performance. However, most of these methods solely focus on domain-level OT alignment by leveraging the geometry of domains for domain-invariant features based on the global embeddings of images. However, global representations of images may destroy image structure, leading to the loss of local details that offer category-discriminative information. This study proposes an end-to-end Deep Hierarchical Optimal Transport method (DeepHOT), which aims to learn both domain-invariant and category-discriminative representations by mining hierarchical structural relations among domains. The main idea is to incorporate a domain-level OT and image-level OT into a unified OT framework, hierarchical optimal transport, to model the underlying geometry in both domain space and image space. In DeepHOT framework, an image-level OT serves as the ground distance metric for the domain-level OT, leading to the hierarchical structural distance. Compared with the ground distance of the conventional domain-level OT, the image-level OT captures structural associations among local regions of images that are beneficial to classification. In this way, DeepHOT, a unified OT framework, not only aligns domains by domain-level OT, but also enhances the discriminative power through image-level OT. Moreover, to overcome the limitation of high computational complexity, we propose a robust and efficient implementation of DeepHOT by approximating origin OT with sliced Wasserstein distance in image-level OT and accomplishing the mini-batch unbalanced domain-level OT. Extensive experiments show the superiority of DeepHOT in several benchmark datasets. The code is released on \url{https://github.com/Innse/DeepHOT}.
\end{abstract}


\section{Introduction}
\IEEEPARstart{I}{n} the context of traditional machine learning, one generally assumes that the training and test data should be independent and identically distributed. However, in many practical vision problems, various factors, such as different acquisition conditions, changing lighting conditions, presence of background, can break this assumption and cause discrepancies in data distribution (a.k.a.  domain shift).  Domain adaptation methods aim to mitigate this issue by finding a shared representation among domains where a classifier can be learned and used for target domain~\cite{7586038}. In this work, we focus on unsupervised domain adaptation (UDA), where no label is available in the target domain.

Fruitful works~\cite{yu2023classification,mei2023denkd, wang2022cluster,wang2022probability,bai2021hierarchical} have been done for unsupervised domain adaptation, where the core idea boils down to aligning distributions across domains. Among others, optimal-transport-based methods have shown superiority in various unsupervised domain adaptation tasks by leveraging the underlying geometric structure of domains in domain-level OT alignment ~\cite{7586038,courty2017joint}. They aim to achieve the overall minimum discrepancy between domains by finding optimal matching flow from source and target domains based on the ground distance of pairwise global representation of images ~\cite{xu2020reliable,li2020enhanced,le2021lamda,damodaran2018deepjdot,courty2017joint}. However, these global representations of images destroy image structures, resulting in the loss of local features. Local features offer essential and transferable information that contributes to the discriminative power~\cite{zhang2020deepemd}. As a result, despite the decreased domain discrepancies, the discriminative power of image features across categories is degraded as well.

In this work, to learn domain-invariant yet category-discriminative representations, we propose a Deep Hierarchical Optimal Transport framework (DeepHOT) for unsupervised domain adaptation by incorporating the domain-level OT and the image-level OT into a unified OT framework to mine the hierarchical structural correlations at both domain-level and image-level. Compared with the conventional domain-level OT approaches for UDA, the ground distance of domain-level OT in the proposed DeepHOT is also an OT at image-level, instead of the distance between global representation of images, resulting in a nested OT framework. The benefit of image-level OT is that it is capable of capturing correspondences between local regions of two images for enhancing discriminative features. Specifically, an image would be divided into a set of local patches, which then allows the patch-to-patch correlations to be encoded into the distance of pairwise images by image-level OT. Subsequently, by employing this distance as the ground distance of domain-level OT, DeepHOT improves the domain-level OT's ability to discriminate between different categories. As a result, in a unified manner, not only domains can be aligned by domain-level OT, but the discriminative power is enhanced by image-level OT.

However, despite the performance superiority, optimal-transport-based methods have the drawback of high computational complexity. To this end, we propose a robust and efficient implementation of the DeepHOT framework. The key approaches to address this issue are three-fold. First, for image-level OT, sliced Wasserstein distance (SWD)~\cite{lee2019sliced} is applied to compute the discrepancy between pairwise images, which has been proved to be more efficient than Earth Mover Distance in time complexity. Second, for domain-level OT, we compute the domain distance over mini-batches and use the averaged result as a proxy for the original problem. And then, time complexity would be considerably reduced if using a small mini-batch. Third, it is difficult for mini-batch OT methods to converge to the true transport plan when the batch size decreases. Therefore, we exploit mini-batch unbalanced optimal transport (UOT)~\cite{fatras2021unbalanced} for domain-level OT, which can provide a more robust solution to the mini-batch sampling.

The contributions of this work are summarized as follows: (1) We propose to learn domain-invariant yet category-discriminative representations for UDA in a unified OT-based and end-to-end manner by a unified deep hierarchical optimal transport method. To the best of our knowledge, it is the first deep hierarchical optimal transport method for UDA. (2) We propose a robust and efficient implementation of the DeepHOT framework, making it possible to apply it for a large-scale dataset in practice. (3) Extensive experiments on four benchmark domain adaptation datasets show consistent improvement over state-of-the-art methods.

\section{Related Work}
\subsection{Unsupervised Domain Adaptation}
Unsupervised domain adaptation aims to learn a model on the labeled source domain that can adapt the target domain in which labels are unavailable. Most existing works can be broadly classified into two categories: discrepancy-based domain alignment and adversarial learning.

\textbf{Discrepancy-based domain alignment.} This kind of methods~\cite{tzeng2014deep,yan2017mind} tend to explicitly minimize a divergence that measures the gap between the source and target domains. A typical discrepancy metric, Maximum Mean Discrepancy (MMD)~\cite{tzeng2014deep}, has been extensively applied to domain adaptation tasks and ~\cite{gretton2012optimal} further proposed multi-kernels MMD. Moreover, Weighted-MMD (WDAN) ~\cite{yan2017mind} integrated class prior with original MMD by class-specific auxiliary weights. Another measure that can be utilized for domain alignment is covariance,  leading to a method Deep Correlation Alignment (Deep CORAL)~\cite{sun2016deep}. These methods only consider aligning the marginal distribution, and however, Deep Transfer Network (DTN)~\cite{zhang2015deep} proposed a conditional alignment method using pseudo labels of target domain. ~\cite{wang2020transfer} learned a joint distribution by dynamically quantifying the relative importance of marginal and class-conditional distribution. Similarly, ~\cite{tian2022unsupervised} dynamically aligned both the feature and label spaces. ~\cite{zhou2023adaptive} employed a teacher-student framework for source and target domains and then aligned their prediction. Along with jointly adapting the marginal and conditional distributions, CKET~\cite{meng2020coupled} further attempted to transfer discriminative information by enforcing the structure consistency between the original feature space and the latent feature space.The same team further underscored the significance of fine-grained cluster structures in the source and target domains~\cite{meng2022exploring} to enhance discriminative power. In recent developments, they have proposed applying structural risk minimization not only at the domain-level but also at the class-level to enhance discrimination~\cite{meng2022dual}. 
Another branch is based on optimal transport which is illustrated in next subsection.

\textbf{Adversarial learning.} Another critical type of methods rely on adversarial training, where the core idea is to maximally confuse domain classifier so as to learn domain-invariant representation~\cite{mei2023automatic}.  Based on this idea,  Domain Adversarial Neural Network (DANN) adversarially~\cite{ganin2016domain} learned a feature extractor and a domain discriminator, and ~\cite{tzeng2017adversarial} had a similar solution for UDA by simply fooling a GAN’s discriminator.  In addition, there are some adversarial image generation methods~\cite{liu2017unsupervised,sankaranarayanan2018generate} that impose a class-conditioning or cycle consistency constraint to capture discriminative embeddings.

\begin{figure*}[t]
\centering
\includegraphics[scale=0.55]{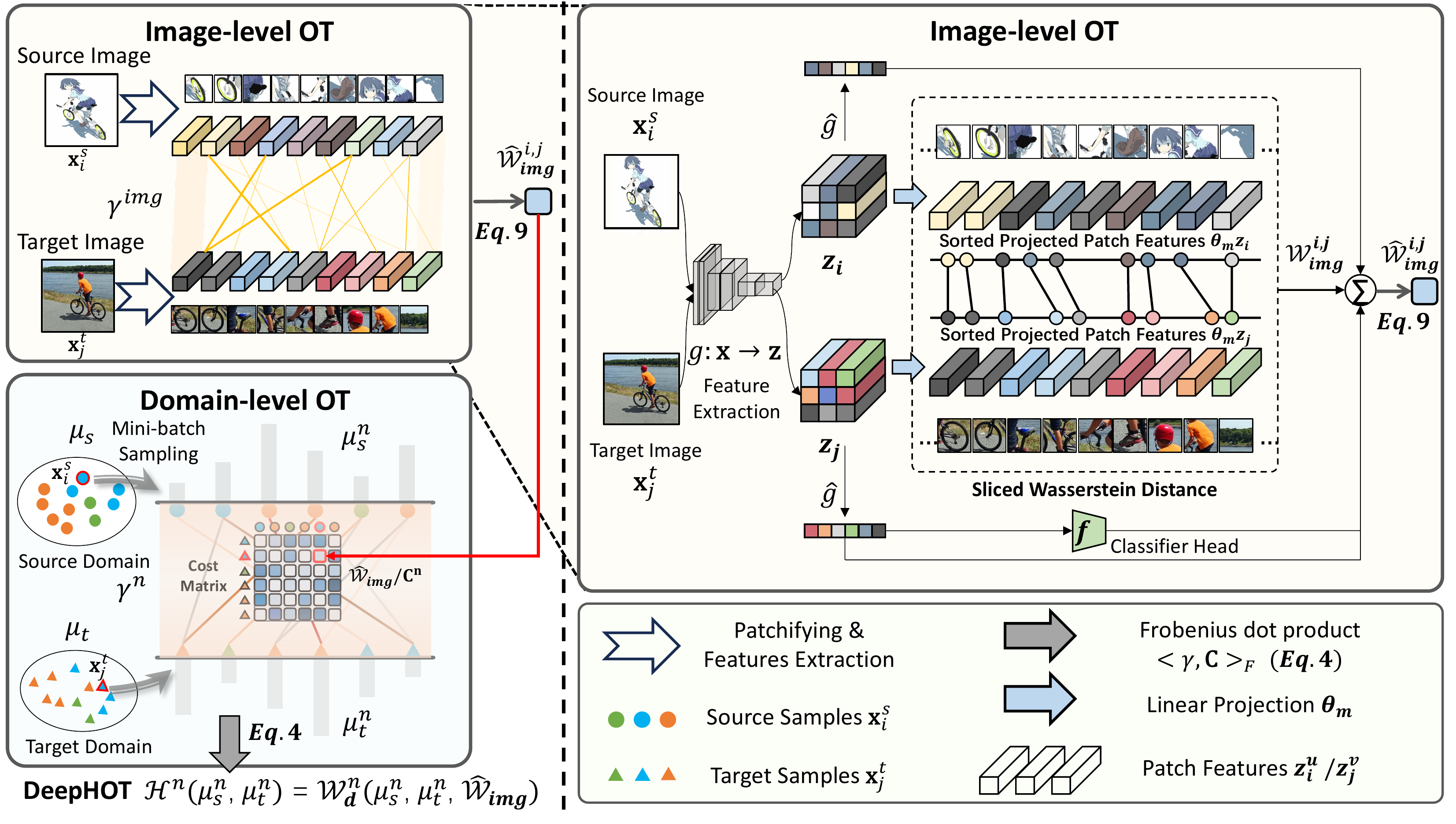}
\caption{Overview architecture of DeepHOT. It consists of domain-level OT and an image-level OT,  where the output of latter is used as the term of cost matrix $\textbf{C}^n$ for the former. And then the distance $\mathcal{H}^n(\mu_s^n, \mu_t^n)$ between $\mu_s^n$ and $\mu_t^n$ domains can be computed via domain-level OT. Colors refer to different classes in domain-level OT and different patches in image-level OT. }
\label{fig:overview}
\end{figure*}

Note that our method DeepHOT belongs to one of discrepancy-based domain alignment methods that minimize the matching cost by optimal transportation to align source and target domains.
\subsection{Optimal Transport for Domain Adaptation}
Optimal Transport (OT) theory has been theoretically proven to be competitive in the context of domain adaptation~\cite{redko2017theoretical,xu2022few}, and ~\cite{7586038} first introduced OT into unsupervised domain adaptation. Then OT-reglab~\cite{courty2014domain} and JDOT~\cite{courty2017joint} further took label information into account where they assumed joint feature/label distribution of source and target domains could be estimated by a non-linear transformation. DEEPJDOT~\cite{damodaran2018deepjdot} extended JDOT into a deep learning-based version, and JUMBOT~\cite{fatras2021unbalanced} alleviated the sampling effect of mini-batch optimal transport which allowed optimal transport for large-scale datasets in an end-to-end way. Recently, TIDOT~\cite{nguyen2021tidot} has jointly mitigated both data shift and label shift by teacher-student framework with optimal transport, and LAMDA~\cite{le2021lamda} has explicitly defined the label shift and minimized it by optimal transportation on the latent space. COOK~\cite{pmlr-v180-nguyen22c} further incorporated the label information into the transformation of OT from source to target domains, leading to the transportation between class-conditional source distribution and a distribution over all the target samples. COT~\cite{liu2023cot} established an OT mapping between learnable clustering centers in the source and target domains in an end-to-end manner. MOT~\cite{luo2023mot} extends OT to a more real-world scenario that the label space typically undergoes changes when the distribution shifts by incorporating a masking strategy for OT between domains. The main difference between the proposed DeepHOT method and the related work is that DeepHOT is a generalized hierarchical optimal transport formulation consisting of domain-level OT and image-level OT for domain adaptation, while the related work such as JUMBOT~\cite{fatras2021unbalanced} involves domain-level OT only.

Another member of optimal transport family is hierarchical optimal transport that is a generalization of OT.  It has been recently applied to different tasks such as multimodal distribution alignment~\cite{lee2019hierarchical},  document representation~\cite{yurochkin2019hierarchical}, semi-supervised learning~\cite{taherkhani2020transporting} and multi-view learning~\cite{luo2020hierarchical}. As far as we know, only one work HOT-DA~\cite{hamri2021hierarchical} has used hierarchical optimal transport for UDA. The differences between our method and HOT-DA are two-fold. First, HOT-DA operated under the assumption that clusters in domains inherently correspond to categories, and thus utilized the knowledge of domain-cluster-level relationships for UDA. However, the granularity of 'cluster' still relies on the global representation of images, resulting in the loss of fine-grained details in an individual image. As a consequence, hard samples, such as the ones with fine-grained discriminative characteristics, become more susceptible to misclassification. In our method, the structural information of images can enhance the discriminative features directly over samples during the adaption process by leveraging domain-image-level relations in hierarchical optimal transport. Second, HOT-DA suffers from another limitation that it cannot be trained on large-scale datasets in an end-to-end manner since it needs to group all samples into clusters offline at first, leading to suboptimal learning of the model. To the best of our knowledge, we propose the first deep learning-based HOT model that can be trained over a mini-batch in an end-to-end way for large-scale UDA tasks.

\section{The Proposed DeepHOT Method}
\subsection{Overview}
Let $\textbf{X}_s=\{ \textbf{x}_i^s \}_{i=1}^{N_s}$ be a set of data drawn from a distribution $\mu_s$ on the source domain, associated with a set of class labels $\textbf{Y}_s=\{ \textbf{y}_i^s \}_{i=1}^{N_s}$, and $\textbf{X}_t=\{ \textbf{x}_i^t \}_{i=1}^{N_t}$ a data set drawn from a distribution $\mu_t$ on the target domain. The labels for the target domain are not available during training. 

In the unsupervised DA problem,  Optimal Transport allows to bring source and target distributions closer by finding optimal transport plan between pairwise samples from source and target domains.  Formally, OT is commonly defined by the Kantorovitch formulation~\cite{kantorovich2006translocation} to search a probabilistic coupling $\gamma \in \Pi(\mu_s, \mu_t) $ between the distribution $\mu_s$ of source domain and $\mu_t$ of target domain:

\begin{equation}
\mathcal{W}_d(\mu_s, \mu_t, c)=\inf_{\gamma \in \Pi(\mu_s, \mu_t)} \int_{\Omega_s \times \Omega_t } c\left(\mathbf{x}^s, \mathbf{x}^t\right) d \gamma\left(\mathbf{x}^s, \mathbf{x}^t \right)
\label{equ:otd_cont}
\end{equation}
where $c\left(\cdot \right)$ is the ground metric that measures the discrepancy between pairwise samples $\mathbf{x}^s$ and $\mathbf{x}^t$. Here, $\Omega_s$ (resp.
 $\Omega_t$) is a compact input measurable space related to source (resp. target) domain, and $\Pi(\mu_s, \mu_t)$ can be understood as the joint probability measure with marginals $\mu_s$ and $\mu_t$.

Here we use discrete optimal transport theory to formulate both domain-level and image-level distances. The empirical distribution $\mu_{s}$ for source domain is formulated as $\mu_{s}=\sum_{i=1}^{N_s} p_{i}^{s} \delta_{\mathbf{x}_{i}^{s}}$ where $\delta_{\mathbf{x}_{i}^s}$ is the Dirac function located at $\mathbf{x}_{i}^s$, and $p_{i}^{s}$ is probability mass associated with $\mathbf{x}_{i}^s$ and belongs to the probability simplex, i.e. $\sum_{i=1}^{N_s} p_{i}^{s}=1$. The empirical distribution $ \mu_{t} $ for target domain can be formulated in the similar way. In this case, we can adapt Kantorovitch OT formulation~\cite{kantorovich2006translocation} to the discrete case.

In a discrete setting,  optimal transport between the empirical distributions is formulated as:
\begin{equation}
\mathcal{W}_d(\mu_s, \mu_t, c) =\min _{\gamma \in \mathcal{B}} <\gamma, \textbf{C}>_F
\label{equ:otd_disc}
\end{equation}
where $\mathcal{B}=\{\gamma \in (\mathbb{R}^+)^{N_s\times N_t} | \gamma\mathbf{1}_{N_t}= \mu_s, \gamma^\mathbf{T}\mathbf{1}_{N_s}=\mu_t \}$, and $\mathbf{1}_d$ is a $d$-dimensional vector of ones. In this case, $\mathcal{B}$ refers to the set of probabilistic couplings between the two empirical distributions. $<\cdot>_F$ is the Frobenius dot product and $\textbf{C} \geq 0$ is the cost matrix, whose term $C(i,j)=c\left(\mathbf{x}^s_i, \mathbf{x}^t_j\right)$ represents the dissimilarity between pairwise samples from source and target domains.  In the general case,  this cost function $c(\cdot)$ is chosen as Euclidean or cosine distance.  

As previously mentioned,  local features of image can help to enhance discriminative power when aligning the marginal distribution across domains.  Here, when decomposing an image into a set of local regions,  we can use optimal transport $\mathcal{W}_{img}$ to measure the distance between samples instead of Euclidean distance, which can be used as the cost function $c(\cdot)$ of problem (\ref{equ:otd_disc}).  As such, the problem (\ref{equ:otd_disc}) becomes a nested OT formulation, i.e.  a general Hierarchical Optimal Transport (HOT) formulation:
\begin{equation}
\label{equ:hot}
\mathcal{H}(\mu_s, \mu_t)=\mathcal{W}_d(\mu_s, \mu_t, \mathcal{W}_{img})
\end{equation}
where correspondences are resolved at two levels: the outer level resolves domain-correspondences while the inner level resolves point-wise correspondences between pairwise samples from source and target domains. Our key insight is depicted in Figure ~\ref{fig:overview}, where image-level OT permits to encode correlations between local image regions to enhance discriminative power for domain alignment that is achieved by domain-level OT. 

Note that domain-level (outer level) OT and image-level (inner level) OT can be any optimal transport algorithm~\cite{chizat2018scaling,frogner2015learning,lee2019sliced,liero2018optimal}. Next we will propose a robust and efficient solution for the generalized DeepHOT framework. In the remainder,  we detail the OT formulation for both domain-level (outer level) and image-level (inner level) in Section ~\ref{sec:domain-level} and ~\ref{sec:image-level}. Finally optimization process and computational complexity of HOT formulation are presented in Section ~\ref{sec:complex}.

\subsection{Domain-Level OT}
\label{sec:domain-level}
The domain-level OT aims to learn the optimal mappings between the samples from source and target domains. To reduce computational complexity, the domain-level OT is trained over mini-batch using deep learning in an end-to-end way. The averaged result among mini-batches is used as a proxy for the original problem. 

To build mini-batch, we select $n$ samples from $\textbf{X}_s$ and $\textbf{X}_t$. Then optimal transport over mini-batch becomes
\begin{equation}
\mathcal{W}_d^n(\mu_s^n, \mu_t^n, \mathcal{W}_{img}) =\min _{\gamma^n \in \mathcal{B}^n} <\gamma^n, \textbf{C}^n>_F
\label{equ:otd_disc_n}
\end{equation}
where $\textbf{C}^n$ is a square matrix of size $n$ and $\mathcal{B}^n=\{\gamma \in (\mathbb{R}^+)^{n\times n} | \gamma\mathbf{1}_{n}= \mu_s^n, \gamma^\mathbf{T}\mathbf{1}_{n}=\mu_t^n \}$.  The term of cost matrix $\textbf{C}^n$ can be computed by optimal transport on image-level $\mathcal{W}_{img}$, which will be illustrated in next subsection.

In practice, however, it was proven that the solution to optimal transport over mini-batch tends to be less sparse, and causes undesired pairings between samples that shouldn't have been matched in true OT~\cite{fatras2021unbalanced}.  A variant~\cite{fatras2021unbalanced} of Unbalanced Optimal Transport (UOT) recently has been theoretically validated to deal with this issue by relaxing marginal constraints. Therefore, we propose to use unbalanced mini-batch optimal transport for the domain-level OT to alleviate the side effect of mini-batch sampling. Then the problem (\ref{equ:otd_disc_n}) becomes
\begin{equation}
\label{equ:otd_uot}
\begin{array}{l}
\mathcal{W}_d^n(\mu_s^n, \mu_t^n, \mathcal{W}_{img}) =\min _{\gamma^n \in \mathcal{B}^n} <\gamma^n, \textbf{C}^n>_F \\
+ \varepsilon K L\left(\gamma^{n} \mid \mu_{s}^{n} \bigotimes \mu_{t}^{n}\right) +\tau\left(D_{\phi}\left(\gamma_{s}^{n} \| \mu_{s}^{n}\right)+D_{\phi}\left(\gamma_{t}^{n} \| \mu_{t}^{n}\right)\right)
\end{array}
\end{equation}
where $\gamma^n$ is the transport plan over mini-batch, $\gamma^n_s$ and $\gamma^n_t$ are plan's marginals. Here, $D_{\phi}$ is Csiszàr divergences, $\tau$ is the marginal penalization and $\varepsilon \geq 0$ is the regularization coefficient. KL is the Kullback-Leible divergence.

This OT formulation (\ref{equ:otd_uot}) is used to replace $\mathcal{W}_d$ in Eq. (\ref{equ:hot}), which naturally proposes a deep learning-based version of HOT method $\mathcal{H}^n(\cdot)$, i.e. Deep \textbf{H}ierarchical \textbf{O}ptimal \textbf{T}ransport (DeepHOT):
\begin{equation}
\label{equ:deephot}
\mathcal{H}^n(\mu_s^n, \mu_t^n)=\mathcal{W}_d^n(\mu_s^n, \mu_t^n, \mathcal{W}_{img})
\end{equation}

\begin{algorithm}[t]
\caption{The DeepHOT Algorithm}
\label{alg:deephot}
\begin{algorithmic}[1]
\Require Source data as $\textbf{X}_s=\{ \textbf{x}_i^s \}_{i=1}^{N_s}$, target data as $\textbf{X}_t=\{ \textbf{x}_i^t \}_{i=1}^{N_t}$. $T$ is set as the number of training iterations, and $n$ refers to the batch-size for training.
\Ensure The optimal parameters of embedding function $g$ and classifier $f$; \\

Randomly initialize $\{\theta_m\}_{m=1}^M$ of M projections.
\For{$t=1$ to $T$}
\State Sample mini-batch from source data $\{ (\textbf{x}_i^s, \textbf{y}_i^s) \}_{i=1}^{n}$ and target data $\{ \textbf{x}_i^t \}_{i=1}^{n}$.
\State Fix $g$, $f$ and $\{\theta_m\}_{m=1}^M$.
\For{$(\textbf{x}_i^s, \textbf{y}_i^s), \textbf{x}_j^t$}
\State // Image-level OT
\State Calculate $\widehat{\mathcal{W}}_{img}^{i,j}((\mathbf{x}^s_i,\textbf{y}_i^s), \mathbf{x}^t_j, c)$ according to Eq. (\ref{equ:swd_comb}).
\State $C(i,j) \gets \widehat{\mathcal{W}}_{img}^{i,j}((\mathbf{x}^s_i,\textbf{y}_i^s), \mathbf{x}^t_j, c)$
\EndFor
\State $\textbf{C}^n=\{C(i,j)\}_{i,j}^{n,n}$
\State Optimize $\gamma^n$ via Sinkhorn-Knopp matrix scaling algorithm~\cite{chizat2018scaling,frogner2015learning}:
\State \ \ \ \ // Domain-level OT
\State \ \ \ \ $\gamma^n \gets \arg\min_{\gamma^n \in \mathcal{B}^n} \mathcal{W}_d^n(\mathbf{x}_s^n, \mathbf{x}_t^n, \widehat{\mathcal{W}}_{img})$ // Eq. (\ref{equ:otd_uot})
\State Fix $\gamma^n$ and calculate the loss of DeepHOT $\mathcal{H}^n(\mathbf{x}_s^n, \mathbf{x}_t^n)$ following Eq. (\ref{equ:deephot_final}) and (\ref{equ:otd_uot}).
\State Calculate the total loss function $\mathcal{L}^n$ according to Eq. (\ref{equ:final_ls}).
\State Update $g$, $f$ and $\{\theta_m\}_{m=1}^M$ by SGD~\cite{sutskever2013importance}.
\EndFor
\end{algorithmic}
\end{algorithm}

\subsection{Image-Level OT}
\label{sec:image-level}
The image-level OT aims to align the image pairs by capturing the correlations among the local regions of images. To approximate OT in a deep learning-based manner, we first formulate an embedding function (e.g. CNN) $g: \mathbf{x} \rightarrow \mathbf{z}$ that maps the input into the latent space $\textbf{Z}$, and the classifier $f: \mathbf{z} \rightarrow \mathbf{y}$ which maps the latent space into the label space $\textbf{Y}$.  

Here latent space can be modelled by any feature layer. In the unsupervised DA task, optimal transport on image-level aims to utilize local information to learn discriminative features.  As such, we use the last convolution layer of a CNN to interpret this latent space that preserves the structural information of feature maps $\mathbf{z} \in \mathbb{R}^{H\times W \times C}$ ($H$ and $W$ denote the spatial size of feature maps and $C$ refers to the channel dimension). And then $\mathbf{z}$ is viewed as a collection of patch features $ \left[\mathbf{z}^{1}, \mathbf{z}^{2}, \ldots, \mathbf{z}^{H W}\right] $ where each element $ \mathbf{z}^{k} \in \mathbb{R}^{C \times 1} (1 \le k \le HW) $ is a vector. The strength of viewing the feature map as a collection of local patch feature vectors is that it allows us to use the optimal transport to mine the structural relationships among local regions of two images, in order to learn the discriminative and transferable information across categories.

In this case, an image $\mathbf{x}_i$ can be represented with a collection of local patch features $[\mathbf{z}_i^{1}, \mathbf{z}_i^{2}, \ldots, \mathbf{z}_i^{HW}]$. Optimal transport can be used to capture the local correspondences between two images which are important for classification. Then we can formulate optimal transport on image-level as 
\begin{equation}
\label{equ:oti_emd}
\begin{aligned}
\mathcal{W}_{img}^{i,j}(\mathbf{x}^s_i, \mathbf{x}^t_j, c)&=\min_{\gamma^{img} \in \Pi(g(\mathbf{x}^s_i), g(\mathbf{x}^t_j))}<\gamma^{img}, \textbf{C}^{img}> \\
&= \min_{\gamma^{img} \in \Pi(\mathbf{z}_i, \mathbf{z}_j)} \sum_u^{HW}\sum_v^{HW}\gamma_{uv}^{img}c(\mathbf{z}_i^{u}, \mathbf{z}_j^{v})
\end{aligned}
\end{equation}
where $c(\mathbf{z}_i^{u}, \mathbf{z}_j^{v})=||\mathbf{z}_i^{u}-\mathbf{z}_j^{v}||^2$.  However, problem (\ref{equ:oti_emd}) can be solved by linear programming, which is not efficient,  especially when combing with domain-level OT.  And thus we introduce a learnable Sliced Wasserstein Distance~\cite{lee2019sliced,luo2020hierarchical} (SWD) to approximate problem (\ref{equ:oti_emd}) for an more efficient OT formulation:
\begin{equation}
\label{equ:oti_swd}
\begin{aligned}
\mathcal{W}_{img}^{i,j}(\mathbf{x}^s_i, \mathbf{x}^t_j, c)&=\frac{1}{M}\sum_{m=1}^M c(sort(\theta_m^\intercal g(\mathbf{x}^s_i)),sort(\theta_m^\intercal g(\mathbf{x}^t_j))) \\
&=\frac{1}{M}\sum_{m=1}^M ||sort(\theta_m^\intercal \mathbf{z}_i)-sort(\theta_m^\intercal \mathbf{z}_j)||^2
\end{aligned}
\end{equation}
where $\{\theta_m\}_{m=1}^M$ contains $M$ linear projectors $\theta_m \in \mathbb{R}^{C\times 1}$ that allow to solve one-dimensional OT problems. In this manner,  we simplify a linear programming problem (\ref{equ:oti_emd}) into equation (\ref{equ:oti_swd}) with the help of a sorting algorithm, where the sorting function sorts the elements of the vector in ascending order, such that $\forall 0 \leq k \leq HW-1, (\theta_m^{\intercal}\mathbf{z}_i)^{(k)} \leq (\theta_m^{\intercal}\mathbf{z}_i)^{(k+1)}$. Then we calculate the Euclidean distance between the sorted vectors and get the averaged result across projectors.

The above term (SWD) together with Euclidean distance are jointly applied as cost function of domain-level OT (i.e. Problem (\ref{equ:otd_uot})) to align marginal distribution. Furthermore, following  ~\cite{courty2017joint,damodaran2018deepjdot,fatras2021unbalanced}, we also use the semantic constraint to handle a conditional distribution shift. Finally, the image-level OT is re-formulated as

\begin{equation}
\label{equ:swd_comb}
\begin{array}{l}
\widehat{\mathcal{W}}_{img}^{i,j}\Big((\mathbf{x}^s_i,\textbf{y}_i^s), \mathbf{x}^t_j, c \Big)\\
=\eta_1 \frac{1}{M}\sum\limits_{m=1}^M \left\Vert sort(\theta_m^\intercal g(\mathbf{x}^s_i))-sort(\theta_m^\intercal g(\mathbf{x}^t_j)) \right\Vert ^2 \\
+ \eta_2 \left\Vert\hat{g}(\mathbf{z}_i)-\hat{g}(\mathbf{z}_j) \right\Vert^2 + \eta_3 \mathcal{L}_{ce}\Big(\textbf{y}_i^s,f(\hat{g}(\mathbf{x}^t_j))\Big)
\end{array}
\end{equation}

where $\hat{g}(\cdot)$ refers to global average pooling of $g(\cdot)$ and $\eta_1$, $\eta_2$, $\eta_3$ are the parameters controlling the tradeoff between these terms. In this way, $\widehat{\mathcal{W}}_{img}(\cdot)$ is used as the cost function of domain-level OT, and thus the term of cost matrix in problem (\ref{equ:otd_uot}) has $C(i,j)=\widehat{\mathcal{W}}_{img}((\mathbf{x}^s_i,\textbf{y}_i^s), \mathbf{x}^t_j, c)$. As a result, DeepHOT model (\ref{equ:deephot}) can be interpreted as
\begin{equation}
\label{equ:deephot_final}
\mathcal{H}^n(\mu_s^n, \mu_t^n)=\mathcal{W}_d^n(\mu_s^n, \mu_t^n, \widehat{\mathcal{W}}_{img})
\end{equation}

To perform classification tasks, the final loss function $\mathcal{L}$ of DeepHOT over mini-batch is integrated with the cross entropy term $\mathcal{L}_{ce}$ on the source data:
\begin{equation}
\label{equ:final_ls}
\mathcal{L}^n =  \mathcal{L}_{ce}(\textbf{X}_s^n, \textbf{Y}_s^n)+ \mathcal{H}^n(\mu_s^n, \mu_t^n)
\end{equation}

\begin{table*}[t]
\centering

\caption{Results on Office-Home (RESNET50). OT refers to OT-based methods, and otherwise Non-OT.}
\begin{tabular}{|c|c|c|c|c|c|c|c|c|c|c|c|c|c|c|}
\toprule
Type& Method & A-C & A-P & A-R & C-A & C-P & C-R & P-A & P-C & P-R & R-A & R-C & R-P & Avg\\
\hline
\multirow{5}{*}{Non-OT}&RESNET50~\cite{he2016identity} & 34.9 & 50.0 & 58.0 & 37.4 & 41.9 & 46.2 & 38.5 & 31.2 & 60.4 & 53.9 & 41.2 & 59.9 & 46.13\\
~&DANN~\cite{ganin2016domain} & 46.2 & 65.2 & 73.0 & 54.0 & 61.0 & 65.2 & 52.0 & 43.6 & 72.0 & 64.7 & 52.3 & 79.2 & 60.70 \\
~&CDAN-E~\cite{long2018conditional} & 52.8 & \underline{71.4} & 76.1 & 59.7 & 70.6 & \underline{71.5} & 59.8 & 50.8 & \underline{77.7} & 71.4 & 58.1 & \underline{83.5} & 66.95\\
~&ALDA~\cite{chen2020adversarial} & 53.7 & 70.1 & \underline{76.4} & 60.2 & \underline{72.6} & \underline{71.5} & 56.8 & 51.9 & 77.1 & 70.2 & 56.3 & 82.1 & 66.58\\
~&SEEBS~\cite{herath2023energy} & \textbf{60.9} & 68.1 & 75.3 & \underline{62.1} & 68.7 & 68.1 & \underline{60.6} & \textbf{60.0} & 75.7 & \underline{72.2} & \textbf{68.2} & 82.0 & 68.49\\
\hline
\multirow{9}{*}{OT}&DEEPJDOT~\cite{damodaran2018deepjdot} & 53.4 & 71.7 & 77.2 & 62.8 & 70.2 & 71.4 & 60.2 & 50.2 & 77.1 & 67.7 & 56.5 & 80.7 & 66.59\\
~&ROT~\cite{balaji2020robust} & 47.2 & 70.8 & 77.6 & 61.3 & 69.9 & 72.0 & 55.4 & 41.4 & 77.6 & 69.9 & 50.4 & 81.5 & 64.58\\
~&JUMBOT~\cite{fatras2021unbalanced} & 55.3 & 75.5 & 80.8 & \underline{65.5} & 74.4 & 74.9 & \textbf{65.4} & 52.7 & 79.3 & \textbf{74.2} & 59.9 & \underline{83.4} & 70.11\\
~&COOK~\cite{pmlr-v180-nguyen22c} & 53.0 & \underline{76.5} & \underline{81.8} & \underline{65.5} & \textbf{80.3} & \textbf{79.2} & \underline{64.5} & 51.8 & \textbf{82.4} & 71.3 & 54.2 & 83.9 & 70.37\\
\cline{2-15}
~&\multirow{2}{*}{\textbf{DeepHOT (ours)}} & \underline{57.0}  & \textbf{77.3} & \textbf{81.9} & \textbf{66.5} & \underline{77.4} & \underline{78.0}& 63.8 & \underline{54.8} & \underline{81.5} & \underline{73.4} & \underline{60.0} & \textbf{84.5} & \multirow{2}{*}{\textbf{71.34}}\\
~& ~ & ±0.2  & ±0.3 & ±0.2 & ±0.1 & ±0.5 & ±0.6& ±0.5 & ±0.3 & ±0.2& ±0.4& ±0.3 & ±0.3 & ~\\
\cline{2-15}
~&DANN~\cite{ganin2016domain} + MMI~\cite{li2021transferable} + COT~\cite{liu2023cot} & 57.6 & 75.2 & \textbf{83.2} & \textbf{67.8} & 76.2 & 75.7 & \textbf{65.4 }& \textbf{56.2} & 82.4 & \textbf{75.1} & 60.7 & 84.7 & 71.68 \\
~&\multirow{2}{*}{\textbf{MMI~\cite{li2021transferable} + DeepHOT (ours)}} & \textbf{57.8}  & \textbf{78.3} & 82.7 & 67.1 & \textbf{76.8} & \textbf{78.2} & 64.6 & 55.6 & \textbf{82.5} & 74.8 & \textbf{61.2} & \textbf{84.8} & \multirow{2}{*}{\textbf{72.03}}\\
~& ~ & ±0.1  & ±0.3 & ±0.2 & ±0.2 & ±0.3 & ±0.5 & ±0.3 & ±0.4 & ±0.1 & ±0.4 & ±0.3 & ±0.1 & ~\\
\bottomrule
\end{tabular}
\label{tab:oh}
\end{table*}

\subsection{Optimization and Computational Complexity}
\label{sec:complex}

We propose an efficient learning algorithm to solve the HOT problem (\ref{equ:deephot_final}) for UDA, which is a new member of Hierarchical Optimal Transport family, combining deep learning-based Unbalanced Optimal Transport on domain-level with Slice Wasserstein Distance on image-level. The details of optimization procedure are presented in Alg.~\ref{alg:deephot}.

In general, OT formulation on each level of HOT model implemented by Wasserstein Distance (or Earth Mover Distance) is solved by linear programming algorithms where the computational complexity is $\mathcal{O}(N^3\log(N))$ and $N=\max(N_s, N_t)$ on domain-level, as well as $\mathcal{O}(K^3\log(K))$ and $K=HW$ on image-level.

Compared with the above regular setting, our DeepHOT implementation has great advantages on computational complexity: (1) The complexity of domain-level OT over mini-batch solved by the generalized Sinkhorn-Knopp matrix scaling algorithm~\cite{chizat2018scaling,frogner2015learning} is $\mathcal{O}(n^2)$ in which $n\ll N$ is the size of mini-batch. It allows to apply HOT model for large scale datasets. For the whole dataset, complexity is $\mathcal{O}(\frac{N}{n}\times n^2)=\mathcal{O}(N\times n)$ which is far smaller than $\mathcal{O}(N^3\log(N))$. (2) On image-level, the complexity of calculating the transport map is $\mathcal{O}(K\log(K))$, where $K$ is small as the spatial size of feature maps (e.g. $K=HW$) down-sampled by CNN is limited. The complexity of calculating $M$ projections of a d-
dimensional probability density functions is $\mathcal{O}(MN^d)$ which dominates the primary complexity.  Hence the overall computational complexity of image-level OT formulation is $\mathcal{O}(MN^d)$.  In this case, we can trade off the computational complexity of image-level formulation by setting the number $M$ of projections.
\section{Experiments}
In this section,  we evaluate our method on four benchmark datasets for UDA tasks, and investigate the efficacy of each component in Ablation Study. Training efficiency is also evaluated and we intuitively validate the effectiveness of proposed method by visualization technique. We rely on the POT package~\cite{flamary2017pot} and PyTorch tools for all experimental implementation.
\subsection{Datasets}

\textbf{Office-Home}~\cite{venkateswara2017deep} contains 15,500 images from four domains: Artistic images (A), ClipArt (C), Product images (P) and Real-World (R), consisting of 65 categories for each domain. We evaluate our method over all 12 adaptation tasks and report results on the test target datasets. 

\textbf{Digits} involves three challenging UDA scenarios following the evaluation protocol of ~\cite{fatras2021unbalanced}: USPS~\cite{hull1994database} to MNIST~\cite{lecun1998gradient} (U$\rightarrow$M), MNIST to M-MNIST~\cite{ganin2016domain} (M$\rightarrow$MM) and SVHN~\cite{netzer2011reading} to MNIST (S$\rightarrow$M). It includes four domains: MNIST with 60,000 images, USPS with 7,291 images, SVHN with 73,212 images and M-MNIST that is colored version of MNIST with 10 classes of digits. 

\textbf{Office-31}~\cite{saenko2010adapting} consists of 4652 images from 31 categories, collected from three domains including Amazon (2817 images), Webcam (795 images) and DSLR (498 images), respectively, where every pair UDA task is evaluated as in ~\cite{huang2022category}. 

\textbf{VisDA}~\cite{peng2017visda} is a large-scale dataset for UDA including two domains: Synthetic and Real-World. It contains 152,397 synthetic images and 55,388 real-world images that share 12 classes. Following the setting of ~\cite{fatras2021unbalanced},  we train all methods on Synthetic and test on VisDA validation set.

\subsection{Experiment Setup}
\label{sec:setup}
\subsubsection{Networks}
For Office-Home, Office-31 and VisDA, we use ResNet-50~\cite{he2016identity} as the generator $g$ and two fully connected layers with dropout as the classifier $f$, where ResNet-50 is pretrained on ImageNet, as done in ~\cite{ganin2016domain,long2018conditional,fatras2021unbalanced}.

For Digits datasets, we employ 9 CNN layers as generator and 1 dense layer for classification, which is the same as previous works ~\cite{damodaran2018deepjdot,fatras2021unbalanced}. Following their settings, we pretrain our model on the source domain for 10 epochs.
\subsubsection{Training}
We present the details of the training setting so as to reproduce the results of the proposed method, including Sampling, Data Augmentation, Optimization and Hyper-parameters.
 
\textbf{Sampling.} For all datasets, all experiments adopt class-balanced sampling on source domain where each class has the same number samples for a mini-batch, and random sampling is used on target domain since labels are unavailable in the target domain. These sampling strategies follow the setting of previous works~\cite{damodaran2018deepjdot,fatras2021unbalanced}.

\textbf{Data Augmentation.} For Office-Home, Office-31 and VisDA,  we use the same data pre-processing where we first resize an image into the size of $256\times 256$ and then randomly crop an area of $224\times 224$. Random translation/mirror and normalization are also applied for training. When testing, we adopt the ten-crop technique~\cite{long2018conditional,fatras2021unbalanced} for robust results. For Digits datasets, the inputs are solely resized and normalized without any other data augmentation. Note that these settings are standard and the same as previous works~\cite{long2018conditional,fatras2021unbalanced} for fair comparisons.

\textbf{Optimization.} Following the settings of ~\cite{fatras2021unbalanced}, for Office-Home, Office-31 and VisDA,  we adopt SGD optimizer with 0.9 momentum and $5e^{-4}$ weight decay for training, and the learning rates are set with strategy~\cite{ganin2016domain} by adjusting $\chi_{p}=\frac{\chi_{0}}{(1+\mu q)^{\nu}}$, where $q$ linearly changes from 0 to 1,  $\chi_{0} = 0.01$, $\mu = 10$, $\nu = 0.75$. Note that the learning rate of classifier $f$ is set to be 10 times that of the generator $g$ as classifier $f$ is trained from scratch. The batch sizes are set as 65, 31 and 72 for  Office-Home, Office-31 and VisDA, respectively, and all experiments are trained for 10,000 iterations. For digits datasets, we apply Adam optimizer with a learning rate of $2e^{-4}$. Due to the limitation of computational complexity, the batch size of DeepHOT is set as 50. We train models for 100 epochs which ensures convergence.

\textbf{Hyper-parameters.} For the domain-level OT, we follow JUMBOT~\cite{fatras2021unbalanced} for the hyper-parameters settings, which include the regularization coefficient $ \epsilon $ and the marginal penalization $ \tau $ in Eq. (5). For the image-level OT, the trade-off hyper-parameters, $ \eta_1, \eta_2, \eta_3 $, are set to scale each loss term to the similar order of magnitude. For the number of projections $ M $ used in Eq. (9), we have conducted parameter sensitivity analysis in the Subsection ~\ref{sec:n_proj}.

The detailed hyper-parameters setting are as follows. The primary hyper-parameters include tradeoff weights $\eta_1$, $\eta_2$, $\eta_3$ of image-level OT in Eq. (9) which are set as follows: $\eta_1=0.001$, $\eta_2=0.001$, $\eta_3=1.0$ for Office-Home,  $\eta_1=0.001$, $\eta_2=0.001$, $\eta_3=0.5$ for Office-31,  $\eta_1=0.01$, $\eta_2=0.001$, $\eta_3=1.0$ for VisDA and $\eta_1=0.1$, $\eta_2=0.1$, $\eta_3=1.0$ for Digits. Another hyper-parameter $\varepsilon$ in Eq. (5) is involved in UOT on domain-level, where $\varepsilon$ is set at $0.01$ for Office-Home, Office-31 and VisDA, and $0.1$ for digits datasets. Moreover, $\tau$ in Eq. (5) is set at $0.5$ for Office-Home and Office-31,  $0.3$ for VisDA and $1.0$ for digits according to ~\cite{fatras2021unbalanced}. $M$ is the number of projections of SWD in Eq.(9).  Here we empirically set it at $128$ for Office-Home and Office-31, $64$ for VisDA and $16$ for digits.

\begin{table*}[ht]
\centering
\caption{Results on Office-31 (RESNET50). OT refers to OT-based methods, and otherwise Non-OT.}
\begin{tabular}{|c|c|c|c|c|c|c|c|c|}
\toprule
Type & Method & A$\rightarrow$W & D$\rightarrow$W & W$\rightarrow$D & A$\rightarrow$D & D$\rightarrow$A & W$\rightarrow$A & Avg \\
\hline
\multirow{6}{*}{Non-OT} & RESNET50~\cite{he2016identity} & 68.4 & 96.7 & 99.3 & 68.9 & 62.5 & 60.7 & 76.08\\
~&DANN~\cite{ganin2016domain} & 82.0 &  96.9 & 99.1 & 79.7 & 68.2 & 67.4 & 82.22\\
~&ADDA~\cite{tzeng2017adversarial} & 86.2 & 96.2 & 98.4 & 77.8 & 69.5 & 68.9 & 82.83\\
~&DWL~\cite{xiao2021dynamic}  & 89.2 & \underline{99.2} & \textbf{100.0} & 91.2 & \underline{73.1} & 69.8 & 87.08\\
~&CaCo~\cite{huang2022category} & 89.7 & 98.4 & \textbf{100.0} & \underline{91.7} & \underline{73.1} & \underline{72.8} & 87.62\\
~&SUDA~\cite{zhang2022spectral} & \underline{90.8} & 98.7 & \textbf{100.0} & 91.2 & 72.2 & 71.4 & 87.38 \\
\hline
\multirow{4}{*}{OT} & DEEPJDOT~\cite{damodaran2018deepjdot}  & 88.9 & 98.5 & 99.6 & 88.2 & 72.1 & 70.1 & 86.23\\
~&JUMBOT~\cite{fatras2021unbalanced} & 93.0±0.4 & 98.7±0.2 & \textbf{100.0}±0 & \textbf{93.0}±0.3 & \underline{73.6}±0.3 & \textbf{75.3}±0.2 & 88.93\\
~&GLOT-DR~\cite{phan2023global} & \textbf{96.2} & \underline{98.9} & \textbf{100.0} & 90.6 & 69.9 & 69.6 & 87.53 \\

\cline{2-9}
~&\textbf{DeepHOT(ours)} & 95.7±0.5 & \textbf{99.3}±0.2 & \textbf{100.0}±0 & 92.8±0.4 & \textbf{74.2}±0.3 & 75.1±0.3 & \textbf{89.52}\\
\bottomrule
\end{tabular}
\label{tab:o31}
\end{table*}

\begin{table}[t]
\centering

\scriptsize
\caption{Results on Digits.}
\begin{tabular}{|c|c|c|c|c|}
\toprule
Method & U$\rightarrow$M & M$\rightarrow$MM & S$\rightarrow$M & Avg \\
\hline
DANN~\cite{ganin2016domain} & 92.2$\pm$ 0.3  & 96.1 $\pm$ 0.6  & 88.7 $\pm$ 1.2 & 92.3 \\
CDAN-E~\cite{long2018conditional} & \textbf{99.2} $\pm$ 0.1 & 95.0 $\pm$ 3.4 & 90.9 $\pm$ 4.8 & 95.0 \\
ALDA~\cite{chen2020adversarial} & 97.0 $\pm$ 1.4 & 96.4 $\pm$ 0.3 & 96.1 $\pm$ 0.1 & 96.5\\
DEEPJDOT~\cite{damodaran2018deepjdot} & 96.4 $\pm$ 0.3 & 91.8 $\pm$ 0.2 & 95.4 $\pm$ 0.1 & 94.5\\
JUMBOT~\cite{fatras2021unbalanced} & 98.2 $\pm$ 0.1 & 97.0 $\pm$ 0.3  & 98.9 $\pm$ 0.1 & 98.0 \\
\textbf{DeepHOT (ours)} & 99.0 $\pm$ 0.2 & \textbf{97.6} $\pm$ 0.3 & \textbf{99.1} $\pm$ 0.1 & \textbf{98.6}\\
\bottomrule
\end{tabular}
\label{tab:digits}
\end{table}

\begin{table}[t]
\centering
\caption{Results on VisDA (RESNET50).}
\begin{tabular}{|c|c|}
\toprule
Method & Accuracy \\
\hline
CDAN-E~\cite{long2018conditional} & 70.1\\
ALDA~\cite{chen2020adversarial} & 70.5\\
DEEPJDOT~\cite{damodaran2018deepjdot} & 68.0\\
ROBUST OT~\cite{balaji2020robust} & 66.3\\
JUMBOT~\cite{fatras2021unbalanced} & 72.5\\
\textbf{DeepHOT (ours)} & \textbf{73.5}$\pm$0.5\\
\bottomrule
\end{tabular}
\label{tab:visda}
\end{table}

\begin{table*}[t]
\centering
\scriptsize
\caption{Ablation study of DeepHOT on Office-Home, where \textbf{(d)} denotes our method, and (a), (b), (c), (e) refer to its variants.}
\begin{tabular}{|l|cc|ccc|c|c|c|c|c|c|c|c|c|c|c|c|c|}
\toprule
\multirow{2}{*}{~}& \multicolumn{2}{|c|}{Domain-level} & \multicolumn{3}{|c|}{Image-level} & \multirow{2}{*}{A$\rightarrow$C}& \multirow{2}{*}{A$\rightarrow$P}& \multirow{2}{*}{A$\rightarrow$R}& \multirow{2}{*}{C$\rightarrow$A}& \multirow{2}{*}{C$\rightarrow$P}& \multirow{2}{*}{C$\rightarrow$R}& \multirow{2}{*}{P$\rightarrow$A}& \multirow{2}{*}{P$\rightarrow$C}& \multirow{2}{*}{P$\rightarrow$R}   & \multirow{2}{*}{R$\rightarrow$A}& \multirow{2}{*}{R$\rightarrow$C}& \multirow{2}{*}{R$\rightarrow$P} & \multirow{2}{*}{Avg} \\
\cline{2-6}
~&EMD & UOT & $l_2$ & $\mathcal{L}_{ce}$ & SWD& ~& ~ &~ & ~ & ~& ~ & ~ & ~ & ~& ~ & ~ & ~ & ~\\
\hline
(a) &~&\checkmark & \checkmark &~&~& 47.4 & 70.0 &75.4 &60.8& 66.6 & 68.1 &59.7& 47.4&76.0 & 71.8 & 54.9&78.6 & 64.7 \\
\hline
(b) &\checkmark &~ & \checkmark &\checkmark &  &53.4&71.7&77.2&62.8& 70.2& 71.4&60.2&50.2 & 77.1 & 67.7 & 56.5 & 80.7 & 66.6 \\
\hline
(c) &~&\checkmark & \checkmark &\checkmark &~& 55.3&75.5& 80.8&65.5& 74.4&74.9&\textbf{65.4}&52.7&79.3 & \textbf{74.2} & 59.9 &83.4 & 70.1 \\
\hline
\textbf{(d)} &~&\checkmark & \checkmark &\checkmark &\checkmark &\textbf{57.0}&\textbf{77.3}&\textbf{81.9}&\textbf{66.5}&\textbf{77.4}&\textbf{78.0}&63.8&\textbf{54.8}&\textbf{81.5} & 73.4 &\textbf{60.0} & \textbf{84.5} & \textbf{71.3}\\
\hline
(e) &~&~& \checkmark &$\checkmark$&$\checkmark$ & 37.2 & 53.2 & 65.7 & 45.1 & 50.9 & 51.6 & 43.4 & 31.2 & 62.8 & 58.3 & 34.9 & 68.3 & 50.2 \\
\bottomrule
\end{tabular}
\label{tab:ablation}
\end{table*}

\begin{figure*}[ht]
\centering
\includegraphics[scale=0.5]{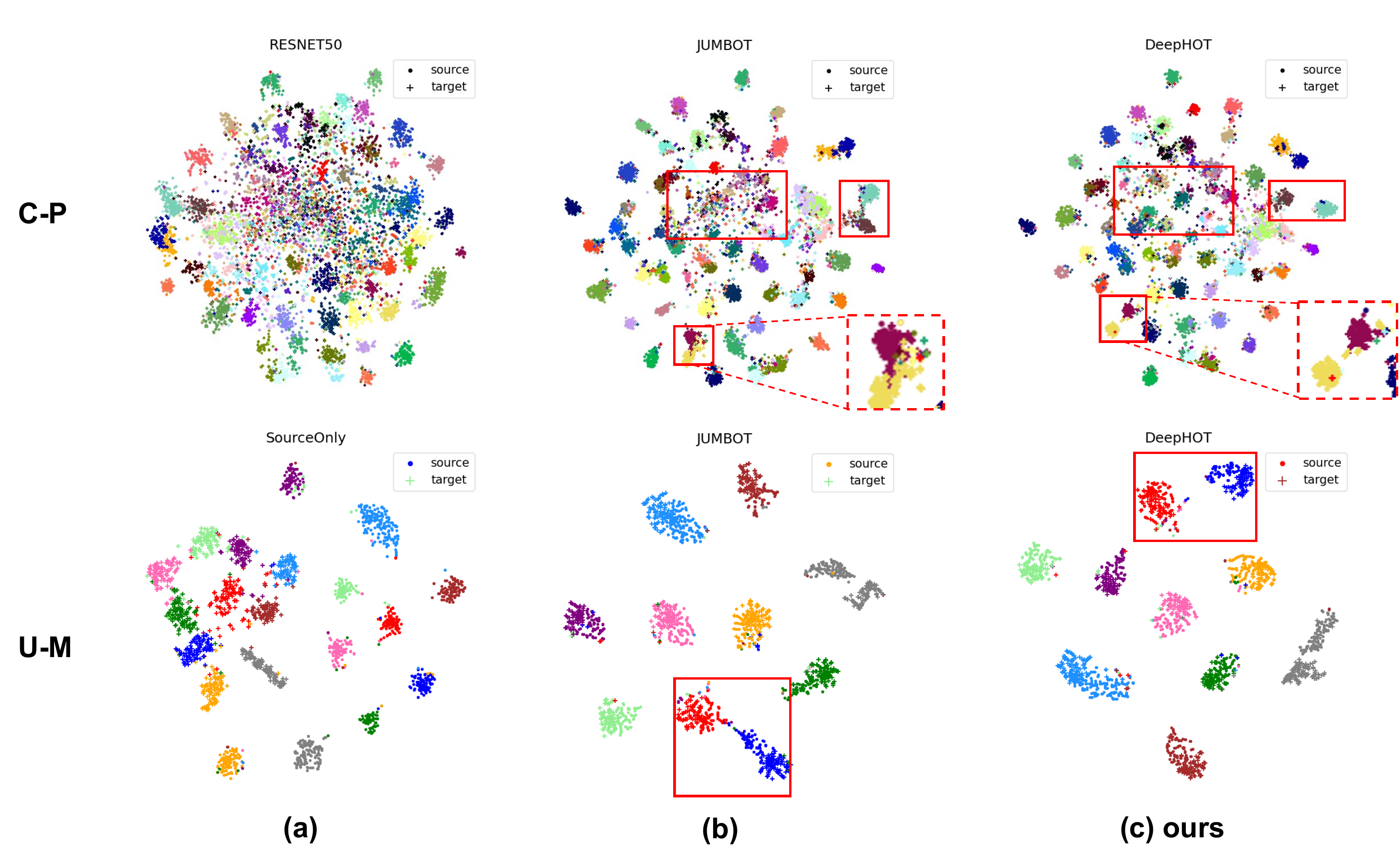}
\caption{The t-SNE visualization.  Representations of task C$\rightarrow$P on Office-Home dataset (65 classes) and task U$\rightarrow$M on Digits dataset (10 classes) are visualized for various methods in (a)-(c), where each color denotes a \textbf{class}.}
\label{fig:tsne}
\end{figure*}
\subsection{Results}
We compare our method against baseline and some state-of-the-art methods, including \textbf{Non-OT} methods (i.e., traditional UDA): DANN~\cite{ganin2016domain}, CDAN-E~\cite{long2018conditional}, ALDA~\cite{chen2020adversarial}, ADDA~\cite{tzeng2017adversarial}, DWL~\cite{xiao2021dynamic}, CaCo~\cite{huang2022category}, SUDA~\cite{zhang2022spectral} and SEEBS~\cite{herath2023energy}, as well as \textbf{OT-based} methods: DEEPJDOT~\cite{damodaran2018deepjdot}, ROT~\cite{balaji2020robust}, JUMBOT~\cite{fatras2021unbalanced}, GLOT-DR~\cite{phan2023global}, COOK~\cite{pmlr-v180-nguyen22c} and COT~\cite{liu2023cot}, that are all based on deep learning with RESNET50~\cite{he2016identity} pretrained on ImageNet for all datasets except digits. Additionally, since COT~\cite{liu2023cot} has not released their code, we just demonstrated their reported results available in their papers with the same settings as ours. Moreover, as COT was trained on the augmented features with MMI~\cite{li2021transferable} based on the model DANN~\cite{ganin2016domain}, we also demonstrated our results on the same features augmented with MMI for fair comparison. We consider \textbf{JUMBOT} as \textbf{baseline}, as it is the most related to ours. The settings on digits of experimental pipeline follow the public implementations of JUMBOT~\cite{fatras2021unbalanced} that is the closest to our method for fair comparisons. We present Accuracy score (in \%) reported in JUMBOT or their own papers unless otherwise stated. We conduct each experiment three times and report the average results and error bars. The error bars for the comparison methods are also shown if they are available in their papers. Despite the varying levels of difficulty among different UDA tasks, we still report average results for reference purposes.

We should claim that the reason for not comparing DeepHOT with HOT-DA~\cite{hamri2021hierarchical} is that HOT-DA can only be evaluated on a subset of the mentioned large-scale datasets due to its high computational complexity. Additionally, HOT-DA lacks explicit details regarding the sampling subsets, making it challenging to reproduce the results. 

\textbf{Results on Office-Home.} As shown in Table ~\ref{tab:oh}, (1) compared with OT-based baseline, JUMBOT, the proposed method gets the better performance on \textbf{10 out of 12} UDA tasks, achieving the best averaged performance. Among others, the highest performance gain arises in C$\rightarrow$R with 3.1\%. (2) Compared with non-OT methods, the proposed approach surpass all of them on 7 out of 12 tasks, which shows the superiority of OT-based methods. (3) Our DeepHOT achieves the best accuracy on 4 of 12 UDA scenarios on the standard ResNet50 features, making it the model with the highest number of best-performing scenarios. Compared with the second best-performing COOK, we get better performance on 8 out of 12 tasks. (4) On the augmented features, the proposed approach surpassed COT on 7 out of 12 tasks, especially on C-R and A-P scenarios with 2.5\% and 3.1\% performance gains. In essence, both COT and our DeepHOT incorporated more fine-grained level alignment in domain adaptation. As a result, they achieve comparable performance and outperform methods that solely focus on domain-level alignment, which further underscores the importance of fine-grained level alignment. These results confirm the superiority of our approach over the other methods, benefiting from discriminative power improvement of image-level OT.

\textbf{Results on Office-31.} Note that results of JUMBOT are reproduced using its official code as they don't report them on their paper. As we can see from Table ~\ref{tab:o31}, (1) compared with OT-based baseline, JUMBOT, our DeepHOT achieves 4 of 6 UDA scenarios, leading to the best averaged performance. (2) Compared with non-OT methods, the proposed method gets the winner on \textbf{all} UDA tasks, which demonstrates the advantage of OT-based methods again. 
(3) Our approach gets the best accuracy on 3 of 6 UDA tasks. 

\textbf{Results on Digits.} Table ~\ref{tab:digits} lists performance of DeepHOT compared with a series of other state-of-the-art methods. As can be seen, DeepHOT achieves a comparable or higher score to the current state-of-the-art methods.It is considerably important that we see increases of the performance among all scenarios compared with JUMBOT that is the most related to our approach. 
The improvement on digits dataset is significant although it enhances not too much, since the accuracy on digits dataset itself is relatively high and is difficult to further improve.

\textbf{Results on VisDA.} Table \ref{tab:visda} gathers results on VisDA dataset, where our method achieves the best accuracy on the validation set with a margin of 1\%. The reported results are convincing since the dataset is large-scale.

From the results mentioned above, the proposed method can achieve the consistent improvement on 4 standard benchmarks of various sizes and difficulties, which provides sufficient empirical evidences for validating the effectiveness of the proposed method.
\begin{figure}[ht]
\centering
\includegraphics[scale=0.31]{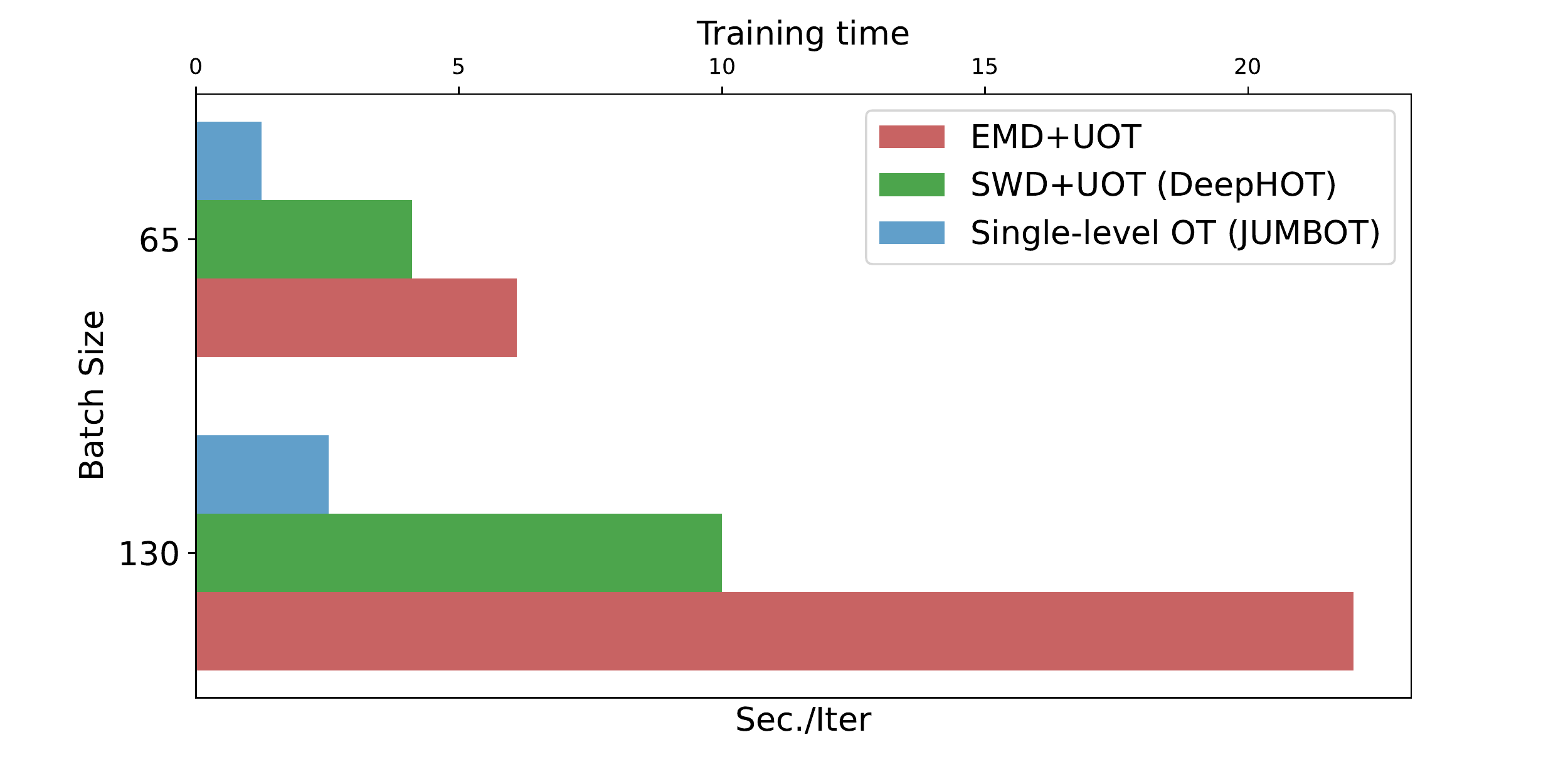}
\caption{Comparison of training time.}
\label{fig:time}
\end{figure}
\subsection{Ablation Study}
In ablation study,  we first validate the efficacy of each component in DeepHOT for both domain-level and image-level, and then we compare the training speed among different implements of OT formulation to prove the efficiency of DeepHOT. We follow the setting of hyper-parameters on Office-Home and each experiment in ablation study adopts the same setting.

\textbf{Component validation.} As shown in Table ~\ref{tab:ablation}, 1) for \textbf{domain-level OT}, we observe that UOT leads to a better performance by comparing the experiments (b) EMD (i.e., origin OT formulation) and (c). 2) For \textbf{image-level OT}, the effectiveness of SWD can be proven from the experiment (c) and (d) in which the performance with SWD further increases. The results of only using image-level OT are shown in the variant (e), achieving the worst performance. The reason is that it ignores the class structure in domains, causing poor classification performance, as what image-level OT does is to minimize the distance between two samples of arbitrary arbitrary class. This also validates the effectiveness of domain-level OT. Based on these consistent improvements, the effectiveness of the proposed hierarchical OT framework composed of domain-level OT and image-level OT can be validated.

\textbf{Training efficiency.} To show the efficiency of SWD on image-level OT, we compare the training time between original OT (a.k.a EMD) and SWD that are combined with domain-level OT (UOT), as well as single-level OT. This experiment runs on the same \textit{GeForce GTX 1080Ti} with 4 GPUs. To ensure a fair comparison, We test the training time on task C$\rightarrow$R of Office-Home from 10th iteration to 100th iteration and get the averaged results. Since the model has not stabilized yet before 10th iteration, performance results are prone to be random and thus we compare results after stable convergence. The number of projections in SWD is fixed at 128. The results are shown in Fig.~\ref{fig:time}. We see that DeepHOT with SWD+UOT (green) achieves a faster speed than that with EMD+UOT (red), which approaches single-level OT method (blue) much more.  It suggests our solution is more efficient among Hierarchical OT-based methods.

\begin{figure*}[t]
\centering
\includegraphics[scale=0.80008]{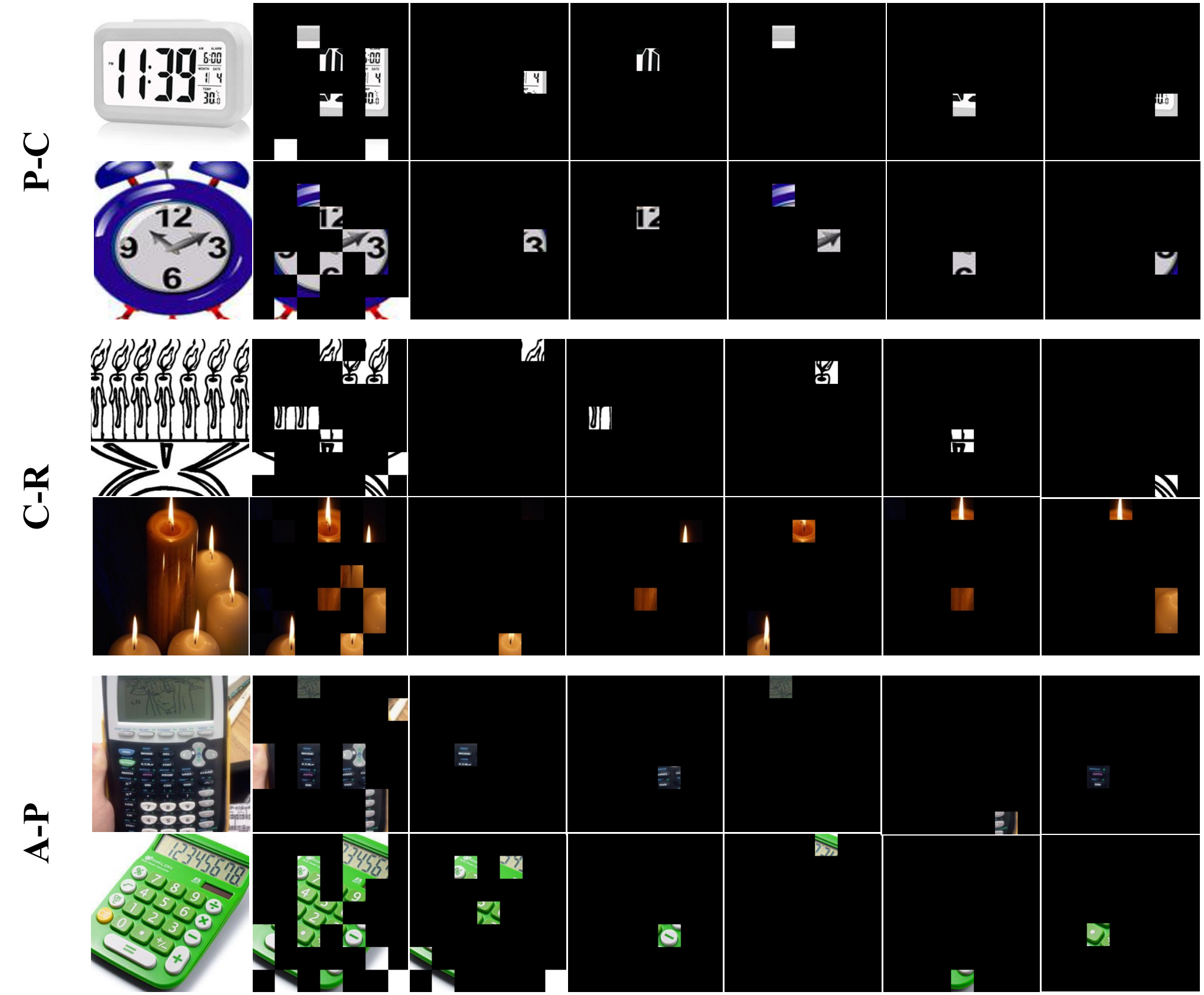}
\caption{Image-level Transport Plan (Consistent Category) on P-C, C-R and A-P adaptation scenarios. }
\label{fig:ot_plan_same}
\end{figure*}

\begin{figure}[ht]
\centering
\includegraphics[scale=0.38]{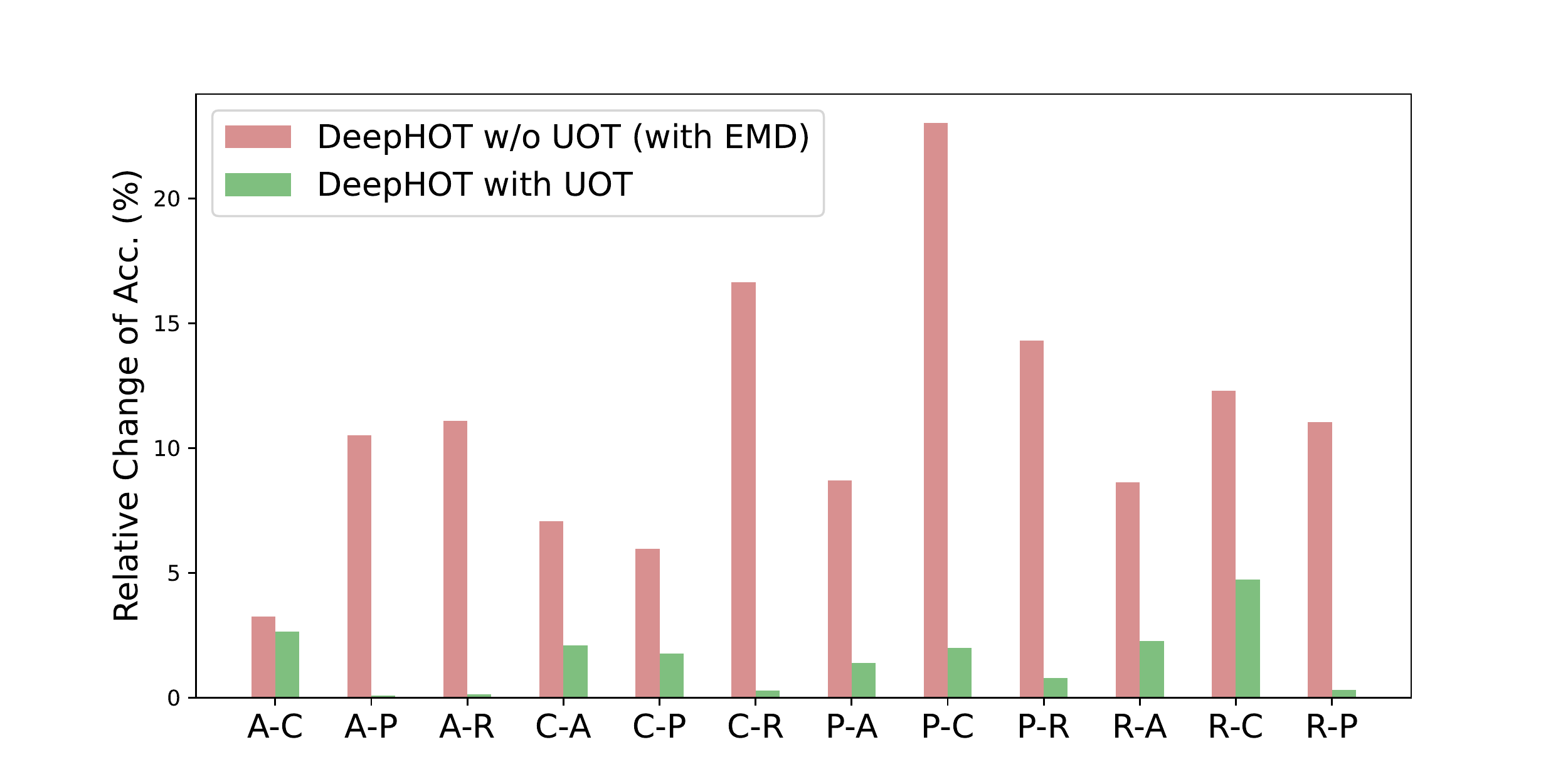}  
\caption{The effect of batch size (from 65 to 130) for DeepHOT with/without UOT.}
\label{fig:bs}
\end{figure}
\begin{figure}[ht]
\centering
\includegraphics[scale=0.27]{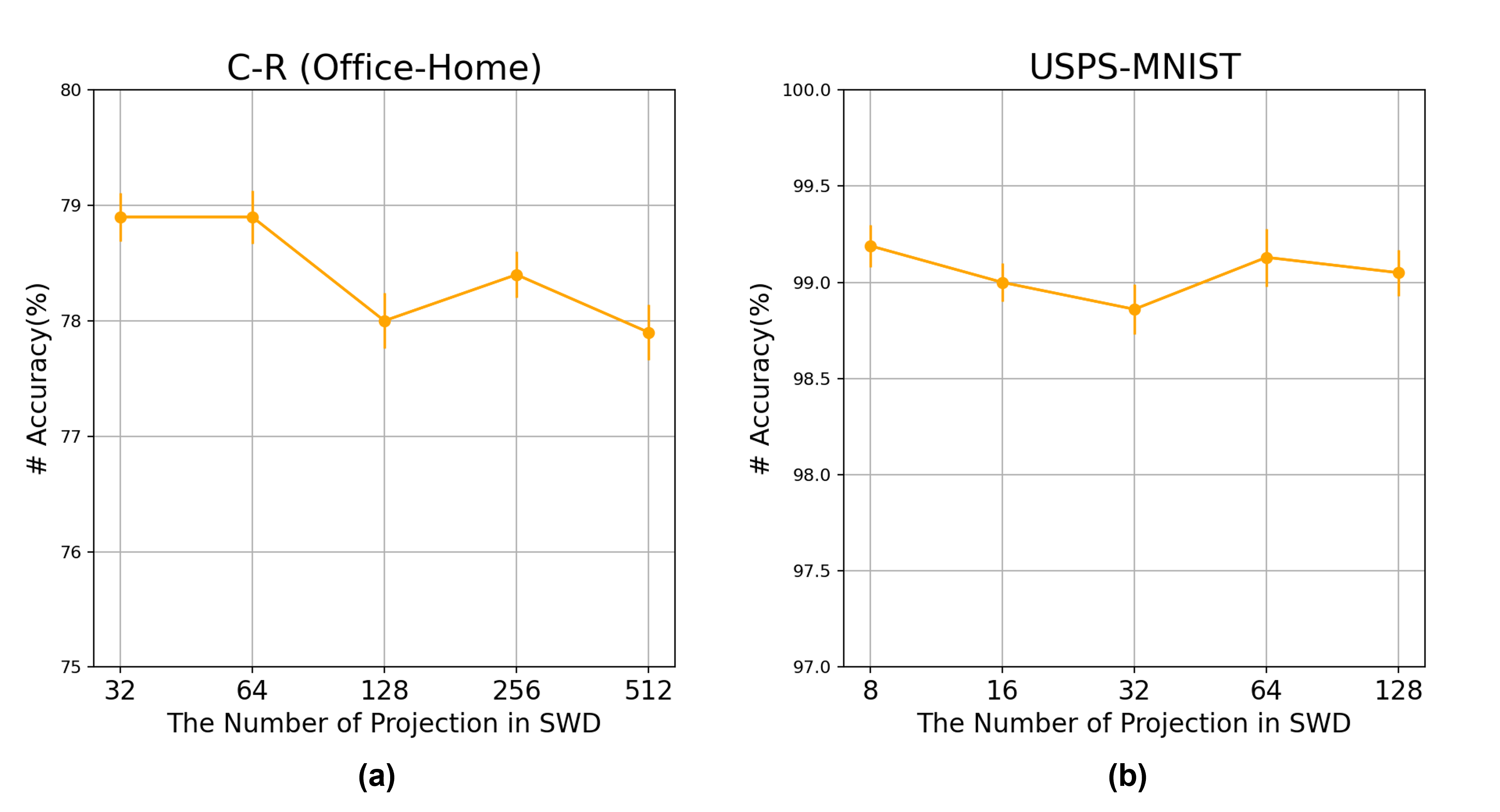}
\caption{The effect of the number of projections in Sliced Wasserstein Distance for tasks (a) C-R (Office-Home), (b) U-M (digits).}
\label{fig:swd_proj}
\end{figure}

\subsection{Parameter Sensitivity Analysis}
We conduct experiments to investigate the parameter sensitivity regarding the batch size for sampling and the number of projections in Sliced Wasserstein Distance.
\label{sec:exp}

\textbf{The Robustness to Batch Size.}
In this part,  we investigate the effect of batch size with/without UOT~\cite{fatras2021unbalanced} for domain-level OT of DeepHOT. Here we fix all hyperparameters and replace UOT with original OT (EMD), compared with DeepHOT. We show the relative change of accuracy from batch size 65 to 130 for DeepHOT with/without UOT in Fig.~\ref{fig:bs}.

We can see that, if using original OT (EMD) instead of UOT in DeepHOT, the model is considerably influenced by the batch size, in which the incremental rate of performance varies from 3\% to 22\% when batch size increases. However, our DeepHOT (with UOT) are robust to the batch size, where all the relative change rates of performances between different batch sizes are less than 5\%. It suggests that our solution based on UOT is robust to mini-batch sampling. Therefore, we can use a small mini-batch to reduce computational complexity.

\textbf{The Number of Projections.}
\label{sec:n_proj}
We conduct sensitivity analysis for the number of projections in Sliced Wasserstein Distance (Eq. (8)) of DeepHOT. Here we vary the number of projections for task C$\rightarrow$R on Office-Home and task U$\rightarrow$M on digits to inspect its effect on the final performances, as shown in Fig.~\ref{fig:swd_proj}. We can see that they slightly change for a series of quantities, suggesting DeepHOT is robust to the number of projections. And thus we can choose a small but appropriate number of projections to reduce computational complexity.

\subsection{Visualization Analysis}

\textbf{t-SNE Visualization.} To present the effectiveness of adaptation process more intuitively, the representations of various methods (no-adaptation, JUMBOT and DeepHOT) are visualized by t-SNE~\cite{van2008visualizing} technique, as shown in Fig.~\ref{fig:tsne}. Here we only show the C$\rightarrow$P scenario on Office-Home dataset and the U$\rightarrow$M scenario on digits dataset, and more cases are presented in supplementary materials. Obviously, the source and target are not aligned without adaptation process, as presented in Fig.~\ref{fig:tsne}(a).  Fig.~\ref{fig:tsne}(b) shows that categories are not separated well, indicating less discrimination. For example, as shown in the red box in the center of Fig.~\ref{fig:tsne} (b) and (c) in C$\rightarrow$P scenario, intuitively the sample points of (b) are more scattered than that of (c), while the boundaries of clusters marked with different colors for different categories in (c) are more clear. Furthermore, there are overlaps in the other red boxes of (b), while the counterparts in (c) are separated. Similarly, as shown in Fig.~\ref{fig:tsne} (b) of U$\rightarrow$M scenario, there are some overlaps on the boundaries between different classes (like red and blue classes) for JUMBOT, while the proposed DeepHOT almost totally distinguishes all the classes. This suggests that our proposal is able to enhance category-discriminative representations when aligning domains.

\begin{figure*}[t]
\centering
\includegraphics[scale=0.80008]{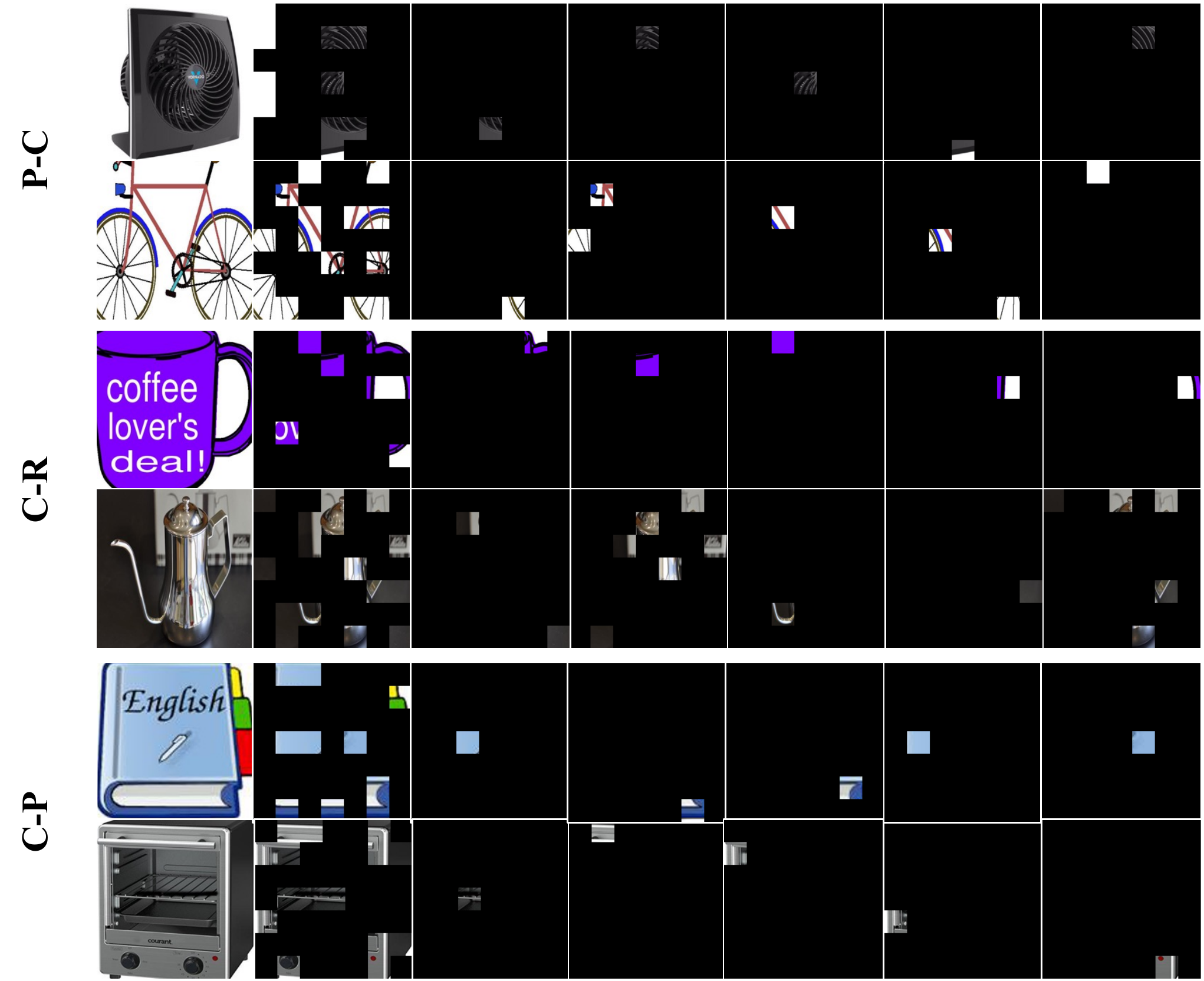}
\caption{Image-level Transport Plan (Different Categories) on P-C, C-R and C-P adaptation scenarios. }
\label{fig:ot_plan_diff}
\end{figure*}

\textbf{Image-level Transport Plan.} To dive into how image-level OT helps the domain adaptation process, we show the transport plan of image-level OT that captures the correspondence of image pairs between source domain and target domains to enhance discrimination. During unsupervised training, the image pairs involve two cases including 'from consistent categories' and 'from different categories' of source and target domains, which are shown in Fig.~\ref{fig:ot_plan_same} and ~\ref{fig:ot_plan_diff}, respectively. In each figure, we show the transport plan for three UDA tasks in which the upper/bottom row is from source/target domain respectively. The first column is original images. The second column is the corresponding image patches of the top 10 indices in descending order of SWD, where we summarize the correlation values of all projectors and get the target patches with maximal correlation value for each source patch (Sometimes there are multiple maximal values, and thus one source patch would correspond to several target patches). The following five columns are the image pairs of the first 5 indices, respectively. More examples are presented in the supplementary materials.

From the image-level transport plan with consistent category in Fig.~\ref{fig:ot_plan_same}, we can see that SWD is capable of transporting the local counterparts from source to target correspondingly, which verified our motivation that local regions of images contribute to enhancing discriminative features, such as the digit of alarm clocks, the flame of candles and the push-button of calculators. Even if transporting image patches between different categories, SWD is still able to transfer some local concepts to help learn in the target domain. For example, as shown in the P-C scenario of Fig.~\ref{fig:ot_plan_diff}, SWD can capture the texture feature of the fan which is similar to the wheel hubs of the bicycle. In the C-R scenario, it also learns the features of handles between Mug and Kettle. For the knowledge-transferring between Notebook and Oven, the image-level OT learns the outline features.

\section{Conclusion}
In this paper, we tackle two important issues for unsupervised domain adaptation. First, how to learn domain-invariant yet category-discriminative representations? We propose the deep hierarchical optimal transport framework to model hierarchical structural distance among domains, which uses the domain-level OT to learn domain-invariant representations, and adopts the image-level OT to capture category-discriminative features. Second, how to build a robust and efficient DeepHOT model? We exploit unbalanced mini-batch optimal transport and sliced Wasserstein distance in domain-level and image-level OTs, respectively. The effectiveness and efficiency of the proposed DeepHOT method are verified by the extensive experiments on four benchmark datasets. It is worth noting DeepHOT is a generalized framework. We will try different implementations of DeepHOT in the future. 

\section*{Acknowledgment}
This work is supported by National Natural Science Foundation of China (Nos. 61976092, 62211530114, and 62006049), Guangdong Basic and Applied Basic Research Foundation (Nos. 2021A1515011317 and 2023A1515010750), Guangdong Key Area Research and Development Plan (No. 2020B1111120001), and International Exchanges 2021 Cost Share (NSFC) Award of The Royal Society (Grant no. IECNSFC211329).

\bibliographystyle{IEEEtran}

\bibliography{ref}

\begin{IEEEbiography}[{\includegraphics[width=1in,height=1.25in,clip,keepaspectratio]{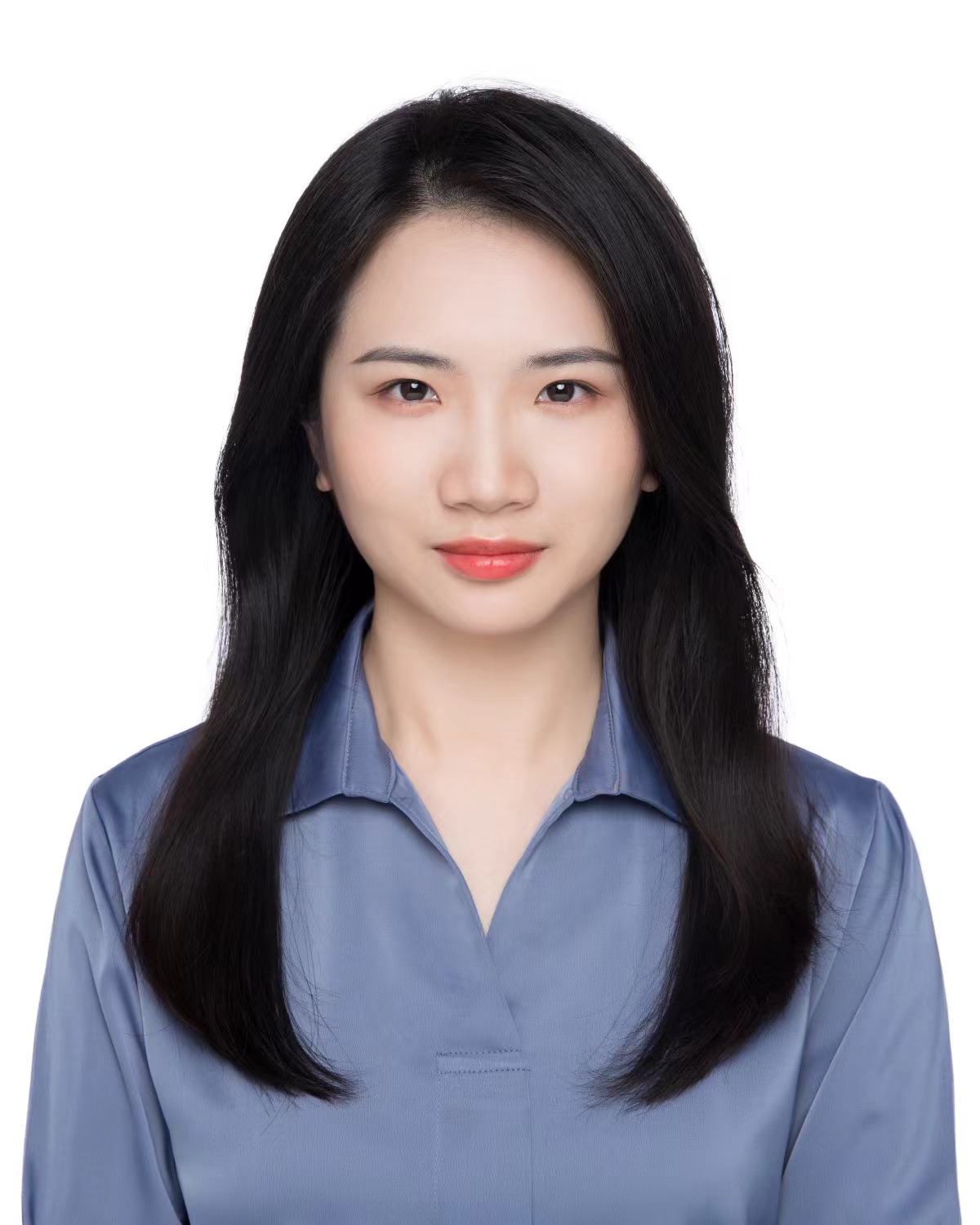}}]{Yingxue Xu}
is currently a Ph.D student in the Department of Computer Science and Engineering, the Hong Kong University of Science and Technology. She received B.Eng and Msc. degree in South China University of Technology. Her main research interests include multimodal learning and medical image analysis.
\end{IEEEbiography}

\begin{IEEEbiography}[{\includegraphics[width=1in,height=1.25in,clip,keepaspectratio]{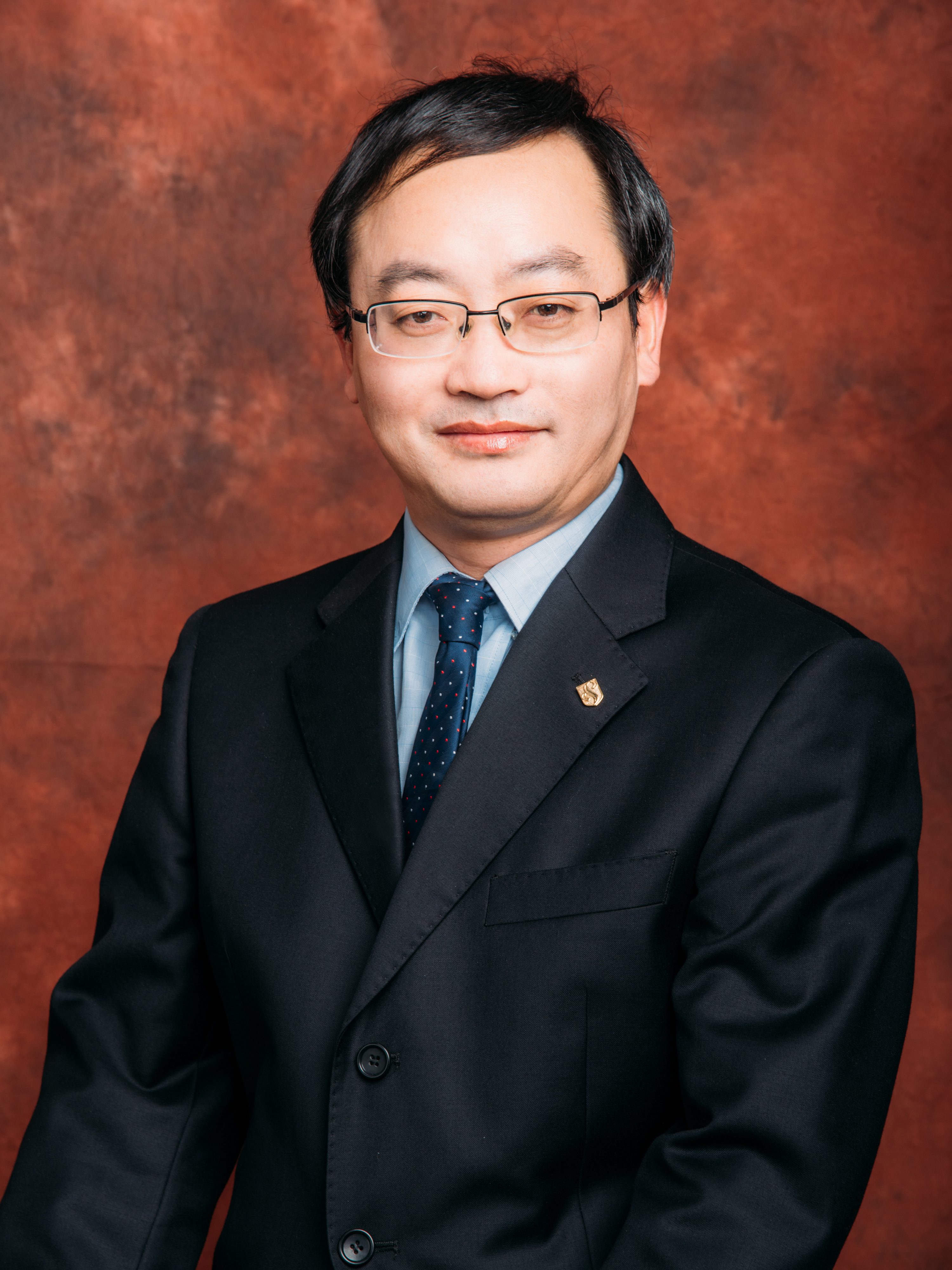}}]{Guihua Wen}
received the Ph.D. degree in Computer Science and Engineering, South China University of Technology, and now is professor, doctoral super-visor at the School of Computer Science and Technology of South China University of Technology. His research area includes Cognitive affective computing, Machine Learning and data mining. He is also professor in chief of the data mining and machine learning laboratory at the South China University of Technology.
\end{IEEEbiography}

\begin{IEEEbiography}[{\includegraphics[width=1in,height=1.25in,clip,keepaspectratio]{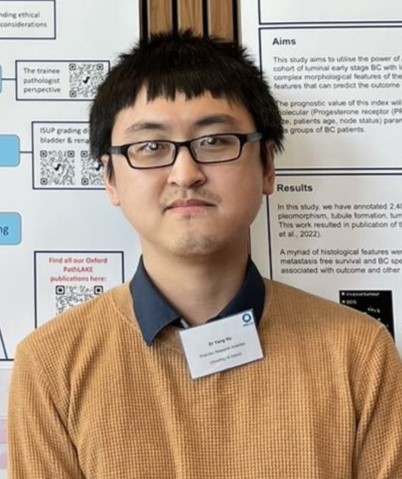}}]{Yang Hu}
(Member, IEEE) received the M.A.Eng. degree from the Kunming University of Science and Technology, Kunming, China, in 2016. He received his PhD degree at the South China University of Technology, Guangzhou, China, in 2020. He was a Research Assistant at University of Southampton, U.K. He is now a Postdoc Research Scientist at University of Oxford, U.K. His research interests include neural networks and deep learning, computer vision, and biomedical information processing.
\end{IEEEbiography}

\begin{IEEEbiography}[{\includegraphics[width=1in,height=1.25in,clip,keepaspectratio]{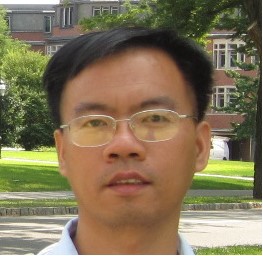}}]{Pei Yang} (Member, IEEE) is an associate professor at South China University of Technology. He received the Bachelor degree from Shanghai Jiaotong University, and the PhD degree from South China University of Technology. He was a research scientist in Arizona State University. His research focuses on statistical machine learning and data mining such as multi-task learning, transfer learning, deep learning, heterogeneous learning, meta learning, few-shot learning, and graph representation learning, with applications on healthcare, computer vision, intelligent transportation, bioinformatics, etc. He has published over 30 research articles on referred journals and top-tier conference proceedings. His paper in ICDM 2016 had been awarded as "Bests of the Conference". He has served on the program committee of conferences such as NeurIPS, KDD, IJCAI, AAAI, TheWebConf, etc.
\end{IEEEbiography}

\end{document}